\documentclass[a4paper]{article}

\usepackage{INTERSPEECH2018}
\usepackage{subfigure}
\usepackage{graphicx}%
\usepackage{float}
\usepackage{enumitem}
\usepackage{ctable} 
\usepackage{multicol, blindtext}

\title{Generating Mandarin and Cantonese F0 Contours\\ with Decision Trees and BLSTMs}
\name{Weidong Yuan, Alan W Black}
\address{
Language Technologies Institute, Carnegie Mellon University, Pittsburgh, USA}
\email{weidongy@andrew.cmu.edu, awb@cs.cmu.edu}

\begin{document}

\maketitle
\begin{abstract}
This paper models the fundamental frequency contours on both Mandarin and Cantonese speech with decision trees and DNNs (deep neural networks). Different kinds of f0 representations and model architectures are tested for decision trees and DNNs. A new model called Additive-BLSTM (additive bidirectional long short term memory) that predicts a base f0 contour and a residual f0 contour with two BLSTMs is proposed. With respect to objective measures of RMSE and correlation, applying tone-dependent trees together with sample normalization and delta feature regularization within decision tree framework performs best. While the new Additive-BLSTM model with delta feature regularization performs even better. Subjective listening tests on both Mandarin and Cantonese comparing Random Forest model (multiple decision trees) and the Additive-BLSTM model were also held and confirmed the advantage of the new model according to the listeners' preference.
\end{abstract}

\section{Introduction}
Decision tree models are widely used for modeling and predicting f0 contours. Variant techniques are applied and proved to be useful. In \cite{teutenberg2008modelling}\cite{wu2008modeling}, Discrete Cosine Transform is introduced as a representation for f0 contours for English and Mandarin respectively. Phrase level and syllable level or even more layers of f0 contours are modeled with different set of features and predicted separately in \cite{teutenberg2008modelling}\cite{wu2008modeling}\cite{sun2002f0}. For improving the smoothness of f0 contours across independently predicted segments, dynamic (delta) feature \cite{yoshimura1999simultaneous} is commonly applied in the models.

In addition to decision tree models, deep neural network models have shown their power for modeling and predicting prosody contours in recent years. Different kinds of DNN architectures are already proposed and very good results are achieved. In \cite{yin2016modeling}, a hierarchical DNN model is proposed predicting different level of f0 contours with different DNNs and lead to a much better result than an HMM baseline model. Bidirectional LSTM is used to predict the prosody contour in \cite{fernandez2014prosody} and outperforms the strong baseline DNN model. In \cite{ronanki2016template}, a template-based LSTM is used for solving the problem of failing to construct good contours through entire utterances in conventional approaches and shows a promising result.

In this paper, we first explore different f0 representations and decision tree model architectures for modeling two speech tone languages Mandarin and Cantonese and the performances are compared. After that, experiments on different traditional deep neural network architectures for predicting syllable level f0 contours are shown. We also propose a simple Additive-BLSTM model which explicitly models lexical information using an additional BLSTM leading to the best performance.

\section{Decision tree}
To make the f0 contours predictable for our models, we split the f0 contours according to the syllables in every utterance. Then 10 values in a syllable f0 contour are subsampled to normalize the duration. So in the models, every sample is a vector consisting of 10 subsampled f0 values. 

\subsection{Features selection} \label{sec:feature}
In decision tree model, f0 contours are modeled and predicted on the syllable level. But different levels of features shown below are used.
\begin{itemize}[leftmargin=2cm]
    \item[Phone level] Vowel; consonant
    \item[Syllable level] Syllable name; duration; phone number in current, previous, next syllable; previous, current, next syllable tone; syllable number from the last, next accented syllable; if the current, previous, next syllable is accented; name of the accent of the current, previous, next syllable; break level after current, previous, next syllable
    \item[Word level] Current, previous, next part of speech, word position in utterance; syllable position in current word
    \item[Phrase level] The current phrase position in the utterance; phrase number in the utterance; syllable number in phrase; stressed syllables from last, next phrase break; number of accented syllables from last, next phrase break; syllable position in phrase
\end{itemize}

\subsection{Model architecture and f0 representation} \label{rep_arch}
In this paper, 5 different f0 representations 4 different architectures are explored for the decision tree model.
\\\\
List for different f0 representations:
\begin{itemize}
\item OriF0 (original f0 vector): 10 subsampled f0 values as a vector are used to represent the f0 contour for every sample (every syllable).
\item DCT: 5 DCT coefficients as a vector are used to represent the f0 contour for every sample.
\item ShapeMS(shape, mean and std): we apply z-score normalization on every sample (f0 vector) in the dataset independently. Then every sample will have its own unique shape vector (the values after normalization), mean and std. In our decision tree models, the mean and std of a sample will be predicted together as a vector but is independently predicted with the shape vector. We call this normalization ``sample normalization'' in the paper.
\item Cross-Delta (f0 representation vector with cross syllable delta): suppose the $t^{th}$ syllable's f0 representation vector in an utterance is $v^{t}$. Cross-Delta for $v^{t}$ is
\begin{equation}
	\Delta v^{t} = [v^{t}-v^{t-1},v^{t+1}-v^{t}]
    \label{eq1}
\end{equation}
where ``[,]'' indicates concatenation of vectors. Then $[v^{t},\Delta v^{t}]$ of each sample is predicted. Note that $\Delta v_i$ here is for the regularization, no backward prediction \cite{zen2009statistical} is needed when estimating. After obtaining the prediction $[\hat{v^{t}},\Delta \hat{v^{t}}]$, $\Delta \hat{v^t}$ will be dropped.
\item In-Delta (f0 representation vector with syllable internal delta): The delta feature is calculated between the f0 values within a sample. Given a f0 representation vector $v^{t} \in R^D$,
\begin{equation}
  \Delta v^{t} = (v^{t}_{2}-v^{t}_{1},v^{t}_{3}-v^{t}_{2},...,v^{t}_{D}-v^{t}_{D-1})
  \label{eq2}
\end{equation}
the same as using Cross-Delta, after making the prediction $[\hat{v^{t}},\Delta \hat{v^{t}}]$, the predicted delta $\Delta \hat{v^{t}}$ will be dropped.
\end{itemize}

List for different model architectures:
\begin{itemize}[leftmargin=1cm]
\item SinDT (single decision tree): single tree predicting vector is used for the prediction (2 trees for ShapeMS (1 for shape vector, 1 for [mean,std] vector)).
\item ToneDT (tone dependent decision tree): tone dependent trees are applied. Each tree is responsible on predicting the f0 vectors belonging to only one specific tone. 
\item PSLevel (phrase level, syllable level additive model): 3 DCT coefficients are used to represent the phrase level contours and predicted by the decision tree\cite{wu2008modeling}. The residual f0 contours will be predicted on the syllable level.
\item ScalarDT (decision trees for predicting scalar): every sample is a vector, instead of predicting the vector, each scalar in the vector is predicted independently by different tree. That is, for the f0 vectors $v \in R^{10}$, the 10 values in the vectors are predicted separately by 10 different trees respectively.
\end{itemize}
Note that different f0 representations are not necessary to be mutually exclusive with each other. The same as different model architectures.

Since DCT is widely used for modeling and predicting the f0 contours \cite{teutenberg2008modelling}\cite{wu2008modeling}, tests on Mandarin and Cantonese speech datasets are done for two classes of model separately: model based on the OriF0 (Table \ref{table:ori_f0}) and model based on the DCT coefficients vector (Table \ref{table:dct_f0}). Some unreasonable combinations of representations and architectures are not shown.

In Table \ref{table:ori_f0},\ref{table:dct_f0}, model (2)(12), model (3) and model (5) show the advantage of ShapeMS (shape, mean and std of sample normalization) representation, In-Delta (syllable internal delta regularization) and ToneDT (tone dependent trees) consistently on both Mandarin and Cantonese speech datasets. Model (8)(11) indicates that DCT coefficients predicted as vector will perform better than predicted separately. However, applying phrase level and syllable level additive model (6)(10) doesn't show improvement here which is surprising. This may be because the speech datasets used here are based on isolated utterances and all have a more standard prosodic phrasing. 

Model (7) using ShapeMS, In-Delta and ToneDT performs best. And applying the random forest model \cite{black2015random} (ignore $30\%$ features and $30\%$ predict predictee coefficients, 20 trees) with model (7) will give us a much better result as shown in Table \ref{table:rf_model}.

\begin{table}
  \caption{Statistics of the OriF0 based models' performance with different f0 representations and architectures on Mandarin and Cantonese speech datasets. ``Syl'' indicates ``Syllable'' and ``Utt'' indicates ``Utterance''. True durations are used.}
  \label{table:ori_f0}
  \centering
  \begin{tabular}{| l | c | c | c | c |}
    \hline
    
    & \multicolumn{2}{|c|}{Mandarin} & \multicolumn{2}{c|}{Cantonese}\\
    \cline{2-3}\cline{4-5}
    \shortstack{Model\\\\\hfil}&
    \shortstack{Syl\\Level\\ \textbf{rmse}\\\textbf{corr}}&
    \shortstack{Utt\\Level\\ \textbf{rmse}\\\textbf{corr}}&
    \shortstack{Syl\\Level\\ \textbf{rmse}\\\textbf{corr}}&
    \shortstack{Utt\\Level\\ \textbf{rmse}\\\textbf{corr}}\\
    \cline{1-5}
	
    \shortstack[l]{(1) OriF0\\ \hspace{0.4cm} SinDT}
    &\shortstack{\\29.214\\0.780}
    &\shortstack{\\33.777\\0.851}
    &\shortstack{\\20.990\\0.674}
    &\shortstack{\\25.946\\0.756}\\
    \hline
    
    \shortstack[l]{(2) ShapeMS\\ \hspace{0.4cm} SinDT}
    &\shortstack{\\28.992\\0.797}
    &\shortstack{\\33.527\\0.854}
    &\shortstack{\\20.923\\0.725}
    &\shortstack{\\25.894\\0.758}\\
    \hline
    
    \shortstack[l]{(3) In-Delta\\ \hspace{0.4cm} SinDT}
    &\shortstack{\\29.094\\0.779}
    &\shortstack{\\33.639\\0.853}
    &\shortstack{\\20.887\\0.683}
    &\shortstack{\\25.829\\0.759}\\
    \hline
    
    \shortstack[l]{(4) Cross-Delta\\ \hspace{0.4cm} SinDT}
    &\shortstack{\\29.494\\0.775}
    &\shortstack{\\34.771\\0.841}
    &\shortstack{\\21.235\\0.674}
    &\shortstack{\\26.032\\0.754}\\
    \hline

	\shortstack[l]{(5) OriF0\\ \hspace{0.4cm} ToneDT}
    &\shortstack{\\29.142\\0.782}
    &\shortstack{\\33.683\\0.852}
    &\shortstack{\\21.033\\0.679}
    &\shortstack{\\26.021\\0.755}\\
    \hline
    
    \shortstack[l]{(6) OriF0\\ \hspace{0.4cm} PSLevel}
    &\shortstack{\\32.267\\0.768}
    &\shortstack{\\36.780\\0.827}
    &\shortstack{\\25.938\\0.663}
    &\shortstack{\\32.080\\0.642}\\
    \hline
    
    \shortstack[l]{(7) ShapeMS\\ \hspace{0.4cm} In-delta\\ \hspace{0.4cm} ToneDT}
    &\textbf{\shortstack{\\28.959\\0.797}}
    &\textbf{\shortstack{\\33.513\\0.854}}
    &\textbf{\shortstack{\\20.814\\0.725}}
    &\textbf{\shortstack{\\25.829\\0.759}}\\
    \hline
  \end{tabular}
\end{table}

\begin{table}
  \caption{Statistics of the DCT based models' performance with different f0 representations and architectures on Mandarin and Cantonese speech datasets. True durations are used.}
  \label{table:dct_f0}
  \centering
  \begin{tabular}{| l | c | c | c | c |}
    \hline
    
    & \multicolumn{2}{|c|}{Mandarin} & \multicolumn{2}{c|}{Cantonese}\\
    \cline{2-3}\cline{4-5}
    \shortstack{Model\\\\\hfil}&
    \shortstack{Syl\\Level\\ \textbf{rmse}\\\textbf{corr}}&
    \shortstack{Utt\\Level\\ \textbf{rmse}\\\textbf{corr}}&
    \shortstack{Syl\\Level\\ \textbf{rmse}\\\textbf{corr}}&
    \shortstack{Utt\\Level\\ \textbf{rmse}\\\textbf{corr}}\\
    \cline{1-5}
    
    \shortstack[l]{(8) DCT\\ \hspace{0.4cm} SinDT}
    &\shortstack{29.145\\0.776}
    &\shortstack{34.546\\0.844}
    &\shortstack{20.967\\0.680}
    &\shortstack{26.079\\0.755}\\
    \hline
    
    \shortstack[l]{(9) DCT\\ \hspace{0.4cm} ToneDT}
    &\shortstack{29.147\\0.778}
    &\shortstack{34.540\\0.844}
    &\shortstack{20.974\\0.682}
    &\shortstack{26.103\\0.755}\\
    \hline

    \shortstack[l]{(10) DCT\\ \hspace{0.5cm} PSLevel}
    &\shortstack{32.221\\0.770}
    &\shortstack{37.573\\0.819}
    &\shortstack{25.894\\0.670}
    &\shortstack{32.644\\0.635}\\
    \hline

    \shortstack[l]{(11) DCT\\ \hspace{0.5cm} ScalarDT}
    &\shortstack{30.728\\0.763}
    &\shortstack{35.940\\0.832}
    &\shortstack{22.770\\0.653}
    &\shortstack{27.878\\0.724}\\
    \hline
    
    \shortstack[l]{(12) ShapeMS\\ \hspace{0.5cm} SinDT}
    &\textbf{\shortstack{29.041\\0.793}}
    &\textbf{\shortstack{34.411\\0.845}}
    &\textbf{\shortstack{20.938\\0.722}}
    &\textbf{\shortstack{26.048\\0.757}}\\
    \hline

    \shortstack[l]{(13) Corss-Delta\\ \hspace{0.5cm} SinDT}
    &\shortstack{29.979\\0.778}
    &\shortstack{35.370\\0.835}
    &\shortstack{21.427\\0.662}
    &\shortstack{26.411\\0.748}\\
    \hline    
  \end{tabular}
\end{table}

\begin{table}
  \caption{Performance of random forest with the best decision tree model}
  \label{table:rf_model}
  \centering
  \begin{tabular}{| c | c | c | c |}
    \hline
    \multicolumn{2}{|c|}{Mandarin} & \multicolumn{2}{c|}{Cantonese}\\
    \cline{1-2}\cline{3-4}
    \shortstack{Syl level\\ \textbf{rmse}\\\textbf{corr}}&
    \shortstack{Utt level\\ \textbf{rmse}\\\textbf{corr}}&
    \shortstack{Syl level\\ \textbf{rmse}\\\textbf{corr}}&
    \shortstack{Utt level\\ \textbf{rmse}\\\textbf{corr}}\\
    \cline{1-4}
    
    \shortstack{\\27.717\\0.814}
    &\shortstack{\\32.049\\0.868}
    &\shortstack{\\20.182\\0.739}
    &\shortstack{\\25.057\\0.774}\\
    \hline
  \end{tabular}
\end{table}

\section{Additive-BLSTM model}
In this section, we investigate the performance of MLP (Multilayer Peceptron), single LSTM (unidirectional LSTM) and single BLSTM which are most commonly used neural network architectures for predicting f0 contours \cite{yin2016modeling}\cite{fernandez2014prosody}\cite{ronanki2016template}. We also propose a new model named Additive-BLSTM which gives us the best result.

\subsection{Features selection}
The features used by different DNN (deep neural network) models here include phone level, syllable level, word level and phrase level features (the same features in Section \ref{sec:feature}). In addition, pretrained word-embeddings for Chinese and Cantonese characters \cite{bojanowski2016enriching} are included as word level feature here.

\subsection{Additive-BLSTM model}
\begin{figure}
  \centering
  \includegraphics[width=0.6\linewidth]{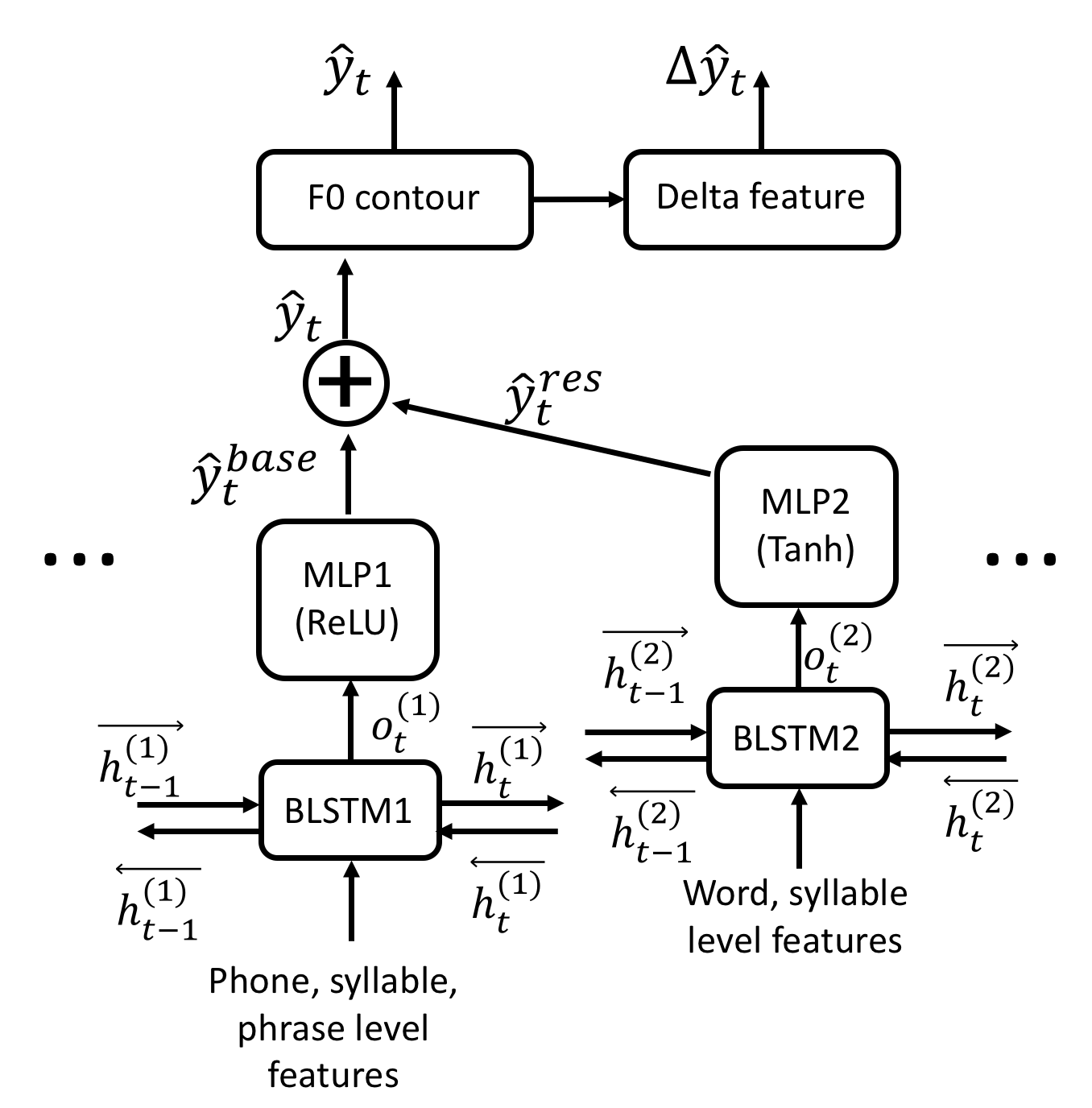}
  \caption{Additive-BLSTM Architecture with delta feature regularization}
  \label{fig:add_lstm}
\end{figure}

Figure \ref{fig:add_lstm} shows the proposed Additive-BLSTM model with delta feature. We use two BLSTMs to handle two sets of features respectively. The first set of features include phone level, syllable level, phrase level features while the second set includes word level, syllable level features. The intuition is that we use one BLSTM (BLSTM1 in Figure \ref{fig:add_lstm}) fed with first set of features together with MLP1 to generate base f0 contour $\hat{y}^{base}_t$. And another BLSTM (BLSTM2 in Figure \ref{fig:add_lstm}) fed with second set of features together with MLP2 is used to capture lexical information from the word sequence of an utterance and help generate residual f0 contour $\hat{y}^{res}_t$. Lexical information is helpful for modeling f0 contours. For example, if the meaning of a word is important in an utterance, the word will be accented which may make its f0 contour rise or fall more quickly. 

Note that MLP1 is a 2-hidden-layer MLP using ReLU activation functions and MLP2 is a 2-hidden-layer MLP with Tanh activation functions. After adding up $\hat{y}^{base}_t$ and $\hat{y}^{res}_t$, we get the predicted contour $\hat{y}_t$. Then delta feature (Cross-Delta or In-Delta) $\Delta \hat{y}_t$ is calculated and used for regularization. 

During training time, mean squared error loss function is used on $[y_t,\Delta y_t]$ and $[\hat{y}_t,\Delta \hat{y}_t]$ ($y_t$ is true f0 contour, $\Delta y_t$ is true delta feature, ``[,]'' indicates concatenation of vectors). During estimation stage, $\Delta \hat{y}_t$ is dropped.

\begin{figure*}[ht]
  \centering
  \includegraphics[width=\textwidth]{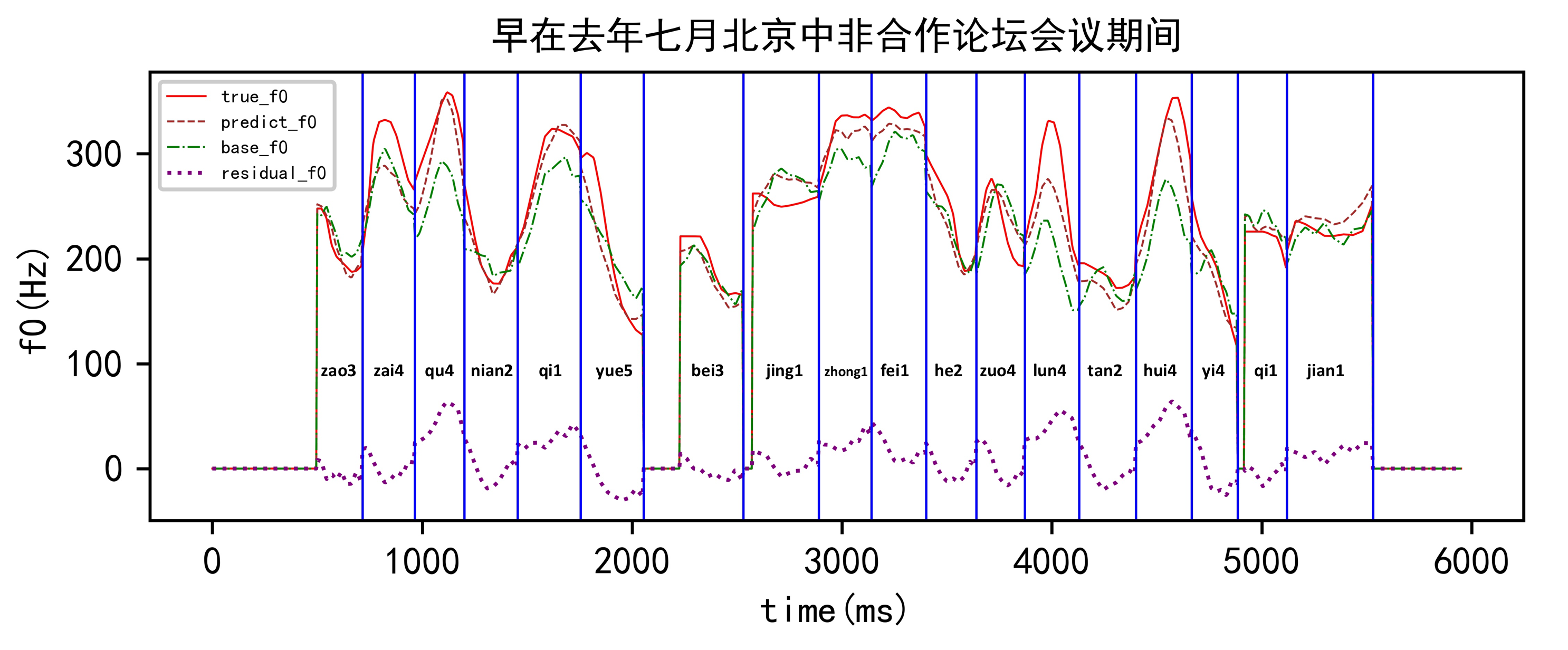}
  \caption{Predicted f0 contour of an example generated by the best Additive-BLSTM model (true durations are used)}
  \label{fig:prediction}
\end{figure*}

\subsection{Performance comparison}

An MLP, a single LSTM, and a single BLSTM are trained as baseline models and are compared with our new model with respect to RMSE and correlation as shown in Table \ref{table:stat_lstm}. For MLP, single LSTM and single BLSTM, features of all the levels are concatenated together as input. 

As shown in Table \ref{table:stat_lstm}, the additive architecture can bring a good improvement to the BLSTM model. And the Additive-BLSTM model with In-Delta performs the best on both Mandarin and Cantonese speech datasets. Figure \ref{fig:prediction} shows the base f0 contour, residual f0 contour and predicted f0 contour for an selected example synthesized by this best Additive-BLSTM model. In the figure, adding the residual f0 contours on the base f0 contours can make the f0 contours rise and fall in a more natural way and also more similar to the natural contours.

\begin{table}[H]
  \caption{Comparison between the performance of MLP, single LSTM, single BLSTM and Additive-BLSTM. As mentioned in section \ref{rep_arch}, Cross-Delta refers to cross syllable delta feature, In-Delta refers to syllable internal delta feature. True durations are used.}
  \label{table:stat_lstm}
  \centering
  \begin{tabular}{| c | c | c | c | c |}
    \hline
    
    & \multicolumn{2}{|c|}{Mandarin} & \multicolumn{2}{c|}{Cantonese}\\
    \cline{2-3}\cline{4-5}
    \multicolumn{1}{|c|}{\shortstack{Model\\\\\hfill}}&
    \shortstack{Syl\\Level \\ \textbf{rmse}\\\textbf{corr}}&
    \shortstack{Utt\\Level \\ \textbf{rmse}\\\textbf{corr}}&
    \shortstack{Syl\\Level \\ \textbf{rmse}\\\textbf{corr}}&
    \shortstack{Utt\\Level \\ \textbf{rmse}\\\textbf{corr}}\\
    \cline{1-5}

    \shortstack{MLP}
    &\shortstack{\\25.721\\0.803}
    &\shortstack{\\30.910\\0.879}
    &\shortstack{\\19.644\\0.715}
    &\shortstack{\\24.821\\0.781}\\
    \hline
    
    \shortstack{Single LSTM}
    &\shortstack{\\24.221 \\0.814}
    &\shortstack{\\29.233 \\0.892}
    &\shortstack{\\19.289\\0.715}
    &\shortstack{\\24.450\\0.787}\\
    \hline
    
    \shortstack{Single BLSTM}
    &\shortstack{\\23.983\\0.818}
    &\shortstack{\\28.925\\0.894}
    &\shortstack{\\19.224\\0.712}
    &\shortstack{\\24.424\\0.789}\\
    \hline
    
    \shortstack{Additive-BLSTM}
    &\shortstack{\\23.467\\0.821}
    &\shortstack{\\28.354\\0.899}
    &\shortstack{\\18.896\\0.723}
    &\shortstack{\\24.046\\0.796}\\
    \hline
    
    \shortstack{Additive-BLSTM\\Cross-Delta}
    &\shortstack{\\23.820\\0.816}
    &\shortstack{\\28.797\\0.895}
    &\shortstack{\\19.328\\0.704}
    &\shortstack{\\24.486\\0.787}\\
    \hline
    
    \shortstack{Additive-BLSTM\\In-Delta}
    &\textbf{\shortstack{\\23.299\\0.828}}
    &\textbf{\shortstack{\\28.266\\0.899}}
    &\textbf{\shortstack{\\18.750\\0.733}}
    &\textbf{\shortstack{\\23.968\\0.797}}\\
    \hline
    
  \end{tabular}
\end{table}

\section{Experiment}
\subsection{Dataset} \label{dataset}
We use two corpora in our experiments for testing the performance of our models. One is the CASIA Mandarin Corpus\cite{zhang2008design} developed for speech synthesis research. 4500 sentences are used for the experiments which includes 76948 Chinese characters (syllables). 3600, 450, 450 sentences are selected randomly as the train data, validation data and test data respectively. Another corpus is CUProsody Cantonese speech dataset. It is a read-speech corpus developed by the DSP and Speech Technology Laboratory of the Chinese University of Hong Kong. It consists of 1,200 newspaper sentences and 1,00 conversational sentences. Only the newspaper sentences are used which include 77164 traditional Chinese characters (syllables). We also split the dataset into 967, 115, 118 sentences as train data, validation data and test data.

\subsection{Experiment setup}
Festival speech synthesis system \cite{black1998festival} is used for the extraction of the F0 contours and most of the features from CASIA Mandarin Corpus and CUProsody. F0 values in an utterance will be split according to syllables and 10 values are subsampled for every syllable. Wagon CART building program in Edinburgh Speech Tools Library\cite{taylor1introduction} is used for building the decision tree models. Besides, Part of speech feature from the raw text is extracted by Stanford CoreNLP toolkit \cite{manning2014stanford}. And FastText pre-trained word embeddings \cite{bojanowski2016enriching} are used in the DNN models.

\subsection{Subjective listening test}
Two subjective listening tests were held. The first test named ``AB Test'' is to compare the sentence pairs synthesized by the best random forest model and the best Additive-BLSTM model respectively. Listeners are asked to select their preference on each sentence pair
played in random order.  The second test named "tone test" is a dictation test to check whether the models generate the correct tones for the syllables. In this test, listeners listen to the sentences with the same pattern ``A teacher wrote A B two words on the blackboard.'' in Mandarin or Cantonese. ``A'' and ``B'' will be replaced by different words. The listeners are asked to write down what they heard for ``A'' ``B'' two words.  We selected two-syllable words that are ambiguous with respect to their tone-type. The carrier phrase is selected to not influence the listener with any semantic context on the word choice.  The tone test is also held for random forest and Additive-BLSTM on both Cantonese and Mandarin.

10 Mandarin speakers and 10 Cantonese speakers participated in the tests for Mandarin speech and Cantonese speech respectively. 20 sentence pairs in AB Test and 20 sentences in tone test were tested for every listener. Table \ref{table:ab_test} and Table \ref{table:tone_test} show the results of two tests. In AB Test, the Additive-BLSTM model is preferred on both Mandarin and Cantonese speech. In tone test, both models have good performance (low error rate) while Additive-BLSTM model is still a little bit better than the random forest model.  Interestingly the errors that listeners made in tone test were sometimes phonetic as well as tonal.

\begin{table}
  \caption{AB test preference result between two models}
  \label{table:ab_test}
  \centering
  \begin{tabular}{| c | c | c | c |}
    \hline
    \shortstack{Language}&
    \shortstack{Additive-\\BLSTM model}&
    \shortstack{Random\\forest model}&
    \shortstack{No\\preference}\\
    \hline
    
    \shortstack{Mandarin}
    &\shortstack{$38.5\%$}
    &\shortstack{$22.0\%$}
    &\shortstack{$39.5\%$}\\
    \hline
    
    \shortstack{Cantonese}
    &\shortstack{$44.5\%$}
    &\shortstack{$31.0\%$}
    &\shortstack{$24.5\%$}\\
    \hline
  \end{tabular}
\end{table}

\begin{table}
  \caption{Tone test tone error rate}
  \label{table:tone_test}
  \centering
  \begin{tabular}{| c | c | c |}
    \hline
    \shortstack{Language}
    &\shortstack{Additive-\\BLSTM model}
    &\shortstack{Random\\forest model}\\
    \hline
    
    \shortstack{Mandarin}
    &\shortstack{$0.85\%$}
    &\shortstack{$3.05\%$}\\
    \hline
    
    \shortstack{Cantonese}
    &\shortstack{$3.61\%$}
    &\shortstack{$4.35\%$}\\
    \hline
  \end{tabular}
\end{table}

\section{Discussion}
Objective evaluation (syllable level, utterance level RMSE and correlation) indicates the advantage of sample normalization, syllable internal delta, tone dependent trees and random forest for decision tree model. However, some techniques like PSLevel model and DCT do not provide improvement on the datasets we use. So more experiments on variable datasets may be needed to explore these techniques comprehensively in the future.

For the BLSTM model, our new additive architecture and syllable internal delta regularization provide good improvement compared with a single BLSTM model. Experiments indicate that using an additional BLSTM fed with word level features like word embeddings and part of speech can capture some lexical information which helps improve the prediction result. But further experiments are still needed to find out what kind of lexical information and how much information are captured by the residual contour.

\section{Conclusions}
In this paper, for modeling the f0 contours of Cantonese and Mandarin speech, multiple f0 representations and model architectures are tested for decision tree and good results are achieved. A new simple Additive-BLSTM is also proposed giving a better f0 contours prediction compared with traditional single BLSTM model. All these improvements are consistent on both Cantonese and Mandarin speech languages. 

In the future, we plan to test our model on more tone languages like Vietnamese, Thai and try to make the model more general for different tone languages.

\bibliographystyle{IEEEtran}

\bibliography{mybib}

\begin{thebibliography}{10}
\providecommand{\url}[1]{#1}
\csname url@samestyle\endcsname
\providecommand{\newblock}{\relax}
\providecommand{\bibinfo}[2]{#2}
\providecommand{\BIBentrySTDinterwordspacing}{\spaceskip=0pt\relax}
\providecommand{\BIBentryALTinterwordstretchfactor}{4}
\providecommand{\BIBentryALTinterwordspacing}{\spaceskip=\fontdimen2\font plus
\BIBentryALTinterwordstretchfactor\fontdimen3\font minus
  \fontdimen4\font\relax}
\providecommand{\BIBforeignlanguage}[2]{{%
\expandafter\ifx\csname l@#1\endcsname\relax
\typeout{** WARNING: IEEEtran.bst: No hyphenation pattern has been}%
\typeout{** loaded for the language `#1'. Using the pattern for}%
\typeout{** the default language instead.}%
\else
\language=\csname l@#1\endcsname
\fi
#2}}
\providecommand{\BIBdecl}{\relax}
\BIBdecl

\bibitem{teutenberg2008modelling}
J.~Teutenberg, C.~Watson, and P.~Riddle, ``Modelling and synthesising f0
  contours with the discrete cosine transform,'' in \emph{Acoustics, Speech and
  Signal Processing, 2008. ICASSP 2008. IEEE International Conference
  on}.\hskip 1em plus 0.5em minus 0.4em\relax IEEE, 2008, pp. 3973--3976.

\bibitem{wu2008modeling}
Z.~Wu, Y.~Qian, F.~K. Soong, and B.~Zhang, ``Modeling and generating tone
  contour with phrase intonation for mandarin chinese speech,'' in
  \emph{Chinese Spoken Language Processing, 2008. ISCSLP'08. 6th International
  Symposium on}.\hskip 1em plus 0.5em minus 0.4em\relax IEEE, 2008, pp. 1--4.

\bibitem{sun2002f0}
X.~Sun, ``F0 generation for speech synthesis using a multi-tier approach,'' in
  \emph{Seventh International Conference on Spoken Language Processing}, 2002.

\bibitem{yoshimura1999simultaneous}
T.~Yoshimura, K.~Tokuda, T.~Masuko, T.~Kobayashi, and T.~Kitamura,
  ``Simultaneous modeling of spectrum, pitch and duration in hmm-based speech
  synthesis,'' in \emph{Sixth European Conference on Speech Communication and
  Technology}, 1999.

\bibitem{yin2016modeling}
X.~Yin, M.~Lei, Y.~Qian, F.~K. Soong, L.~He, Z.-H. Ling, and L.-R. Dai,
  ``Modeling f0 trajectories in hierarchically structured deep neural
  networks,'' \emph{Speech Communication}, vol.~76, pp. 82--92, 2016.

\bibitem{fernandez2014prosody}
R.~Fernandez, A.~Rendel, B.~Ramabhadran, and R.~Hoory, ``Prosody contour
  prediction with long short-term memory, bi-directional, deep recurrent neural
  networks.'' in \emph{Interspeech}, 2014, pp. 2268--2272.

\bibitem{ronanki2016template}
S.~Ronanki, G.~E. Henter, Z.~Wu, and S.~King, ``A template-based approach for
  speech synthesis intonation generation using lstms.'' in \emph{INTERSPEECH},
  2016, pp. 2463--2467.

\bibitem{zen2009statistical}
H.~Zen, K.~Tokuda, and A.~W. Black, ``Statistical parametric speech
  synthesis,'' \emph{Speech Communication}, vol.~51, no.~11, pp. 1039--1064,
  2009.

\bibitem{black2015random}
A.~W. Black and P.~K. Muthukumar, ``Random forests for statistical speech
  synthesis,'' in \emph{Sixteenth Annual Conference of the International Speech
  Communication Association}, 2015.

\bibitem{bojanowski2016enriching}
P.~Bojanowski, E.~Grave, A.~Joulin, and T.~Mikolov, ``Enriching word vectors
  with subword information,'' \emph{arXiv preprint arXiv:1607.04606}, 2016.

\bibitem{zhang2008design}
J.~T. F. L.~M. Zhang and H.~Jia, ``Design of speech corpus for mandarin text to
  speech,'' in \emph{The Blizzard Challenge 2008 workshop}, 2008.

\bibitem{black1998festival}
A.~Black, P.~Taylor, R.~Caley, and R.~Clark, ``The festival speech synthesis
  system,'' 1998.

\bibitem{taylor1introduction}
P.~Taylor, A.~Black, and R.~Caley, ``Introduction to the edinburgh speech
  tools, 1999,'' \emph{Currently available at http://www. cstr. ed. ac.
  uk/projects/speech\_tools/manual-1.2. 0}.

\bibitem{manning2014stanford}
C.~Manning, M.~Surdeanu, J.~Bauer, J.~Finkel, S.~Bethard, and D.~McClosky,
  ``The stanford corenlp natural language processing toolkit,'' in
  \emph{Proceedings of 52nd annual meeting of the association for computational
  linguistics: system demonstrations}, 2014, pp. 55--60.

\end{thebibliography}


\end{document}